# Deep Convolutional Neural Network-Based Autonomous Drone Navigation

K. Amer, M. Samy, M. Shaker and M. ElHelw
Center for Informatics Science
Nile University
Giza, Egypt.
{k.amer, m.serag, melhelw}@nu.edu.eg

## Abstract

*This paper presents a novel approach for aerial drone autonomous navigation along predetermined paths using only visual input form an onboard camera and without reliance on a Global Positioning System (GPS). It is based on using a deep Convolutional Neural Network (CNN) combined with a regressor to output the drone steering commands. Furthermore, multiple auxiliary navigation paths that form a 'navigation envelope' are used for data augmentation to make the system adaptable to real-life deployment scenarios. The approach is suitable for automating drone navigation in applications that exhibit regular trips or visits to same locations such as environmental and desertification monitoring, parcel/aid delivery and drone-based wireless internet delivery. In this case, the proposed algorithm replaces human operators, enhances accuracy of GPS-based map navigation, alleviates problems related to GPS-spoofing and enables navigation in GPS-denied environments. Our system is tested in two scenarios using the Unreal Engine-based AirSim [32] plugin for drone simulation with promising results of average cross track distance less than 1.4 meters and mean waypoints minimum distance of less than 1 meter.*

## 1. Introduction

Due to their low cost and versatility, Unmanned Aerial Vehicles (UAVs), also known as aerial drones or drones for short, are being utilized in many applications [42, 43, 44, 45, 46] such as traffic monitoring, parcel delivery, surveillance and reconnaissance, environmental and desertification monitoring, drone-based wireless internet delivery, drone taxis, to name a few. In many of these applications, the drone regularly travels to the same location(s) to collect data or drop shipments with drone navigation achieved through human operators or autonomously. The latter mode relies on onboard sensors, such as Inertial Measurement Units (IMU) and Global Positioning Systems (GPS) [30, 13] to autonomously navigate the drone along predefined paths. Such path-following approach typically integrates GPS localization with a closed-loop drone navigation control system using the IMU. The feedback from the IMU and GPS enables the drone to travel along certain paths and correct for any drift.

Reliance on GPS-based localization only has many weaknesses, however. Being a radio signal, GPS is susceptible to signal interference from different sources. Some environments such as those surrounded by high rise buildings are GPS-denied due to signal unavailability. Even when the drone is in line-of-sight with GPS satellites, the information in the remote signal can be manipulated by attackers with malicious intents, *e.g.* GPS spoofing [39, 37]. Consequently, GPS-based drone navigation can suffer from sustained drift due to accumulative navigation errors.

In this work, we propose a system for autonomous path following using visual information acquired from a drone-mounted monocular camera. Onboard information has the advantage of being available to be used instantaneously without dependence on external signals and is not likely to be exposed to spoofing attacks. Furthermore, visual data can be used to attain robust navigation since images represent rich source of information with discriminative features that sufficiently describe waypoints along a certain path. The proposed system employs simulator-generated data along with data collected during manually-operated or GPS-based trips to generate a combined dataset. This dataset is subsequently used to train a deep Convolutional Neural Network (CNN) [19, 34] to generate drone steering commands based on observed imagery and achieve autonomous drone navigation. We review some of the related work in Section 2, describe the details of our system in Section 3 and the data collection process in Section 4. Section 5 presents carried out experiments and results whereas Section 6 concludes the paper and points out possible future work directions.

## 2. Related Work

CNNs had achieved remarkable success in the problem of image classification in the ImageNet Large Scale Visual Recognition Competition (ILSVRC) [9] and have since been used in other computer vision tasks such as localization [31], detection [26, 10], and segmentation [2, 27] where their performance in these tasks also outperformed traditional computer vision algorithms.

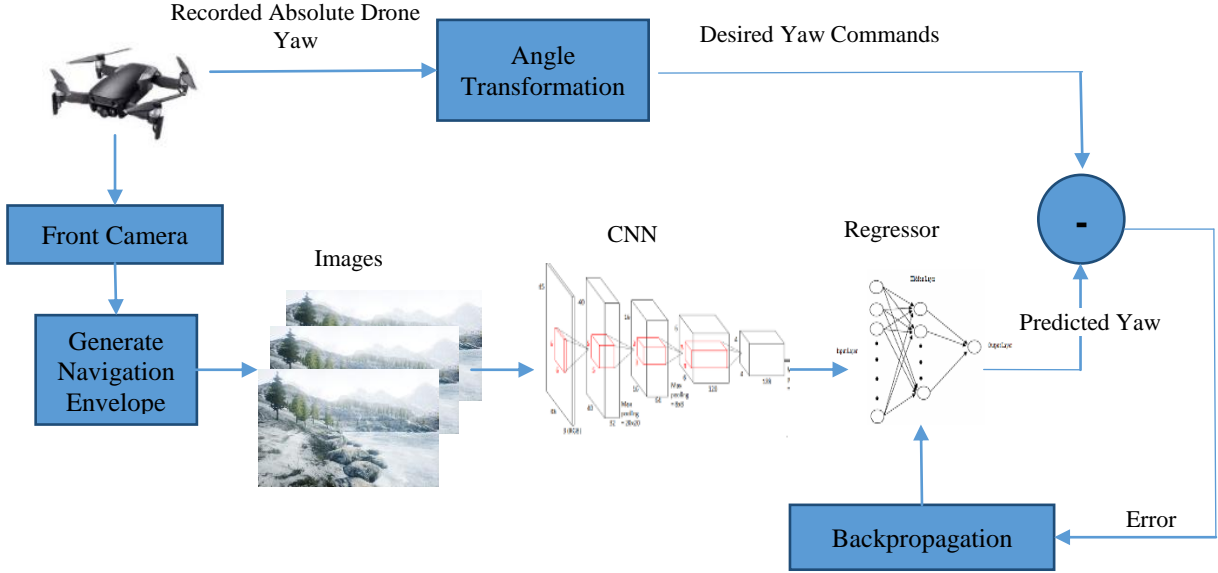

Figure 1: Our proposed system for path following using visual information. A drone is flies on a predetermined path waypoints while recording the current front view and yaw of the drone. To construct a robust system, we control this drone to fly inside an envelope around the waypoints to generate more data. Secondly, the recorded yaw is transformed to relative yaw commands. Finally, we train a neural networks combined of a CNN and a regressor in order to control the drone o the path using only visual information.

Recently, CNNs have been used for achieving autonomous navigation behavior for a variety of mobile platforms such as vehicles, robots and aerial drones. In vehicle navigation based on visual input [3], an end-to-end trained deep CNN is used to map input image from a front facing camera to a steering angle. The network uses visual input to associate straight road segments with small angles and curved segments with a suitable angle to keep the vehicle in the middle of lane. Zhang et. al. [40] combined Reinforcement Learning (RL) with model predictive control to train a deep neural network on drone obstacle avoidance. Zho et. al. [41] proposed an RL-based model that navigates a drone indoors while visually searching for target objects. Their Siamese model takes as input image of the target object and ascene observations representing states. Chaplot et. al. [5] introduced Active Neural Localization (ALM) which predicts a likelihood map for an agent location on a map and uses this information to predict a navigation policy. The main advantage of using RL is it doesn't need manual data labeling. Imitation Learning (IL)- based models represent another approach for autonomous navigation where models learn to mimic human behavior. Ross et. al. [29] are able to produce an onboard drone model to avoid obstacles in a forest using the DAgger algorithm, [28] a widely used IL algorithm. Besides visual features, the trained model depends on different kind of features such as low pass filtered history of previous commands. Kelchtermans et. al. [15] used an GRU recurrent neural network [47] and IL to perform navigation based on a sequence of input images. Giusti et. al. [11] developed a deep learning-based model to control a drone to fly over forest trails where the actions produced by the network are discrete (go right - go straight - go left). Smolyanskiy et. al. [35] used similar model for trail following and added entropy reward to stabilize drone navigation. Kim et. al. [17] trained a CNN model for specific target search in indoor environments. In order to increase model generalization, they augmented the training data and started the training with a pretrained model. Other work in the literature focuses on path following and destination-specific navigation. Brando et. al. [4] developed a method for shoreline following on water banks and similar patterns using Gaussian low-pass filter followed by moving average smoothing on the received input image to control the aerial drone. De Mel et. al. [8] used optical flow [14] between consecutive frames to calculate the relative position of the drone. Nguyen et. al. [24] extended the Funnel Lane theory [6] to control the drone's yaw and height using Kanade-Lucas-Tomasi (KLT) corner features [20, 33]. Having a separate localization phase can help attain more accurate navigation. Amer et. al. [1] designed a global localization system by classifying the district above which the drone is flying. Wang et. al. [38] developed an end-to-end system for calculating odometery based on visual information. Clark et. al. [7] proposed integrating inertial measurements obtained from the onboard IMU unit with visual information from camera to localize a drone in an indoor environment. Kendall et. al. [16] trained a CNN to predict the camera location and orientation directly from a single input image in outdoor scenes based on adjacent landmarks. Melekhov et. al. [21] calculated relative pose from two images by fine tuning a pretrained CNN.

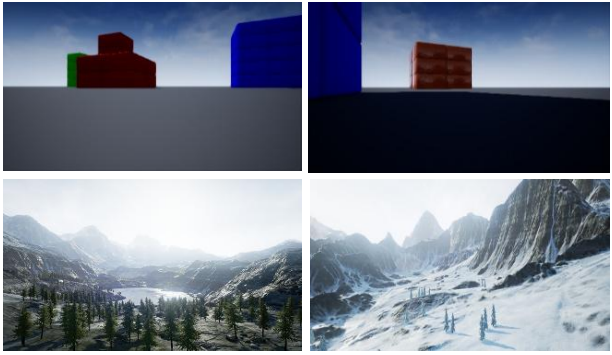

**Figure 2:** Image samples from the simulated environments used in this work. First row from the Blocks environment and second row from the Landscape environment.

In this work, we focus on the drone path following problem where the desired navigation path is specified in terms of waypoints typically defined by selecting geolocations on an offline map of the flight zone. The drone is required navigate in shortest possible path that coincides with all waypoints. The proposed approach trains a deep CNN on footages captured by drone-mounted camera during real and/or simulated trips that approximately follow the desired path. Once trained, the proposed system is subsequently used to autonomously guide the drone along the desired path or complement traditional GPS-based navigation systems for increased accuracy and/or to alleviate GPS-spoofing effects. Typical scenarios for our system include package delivery to regular customers in GPS-denied urban areas, air taxis, environmental and desertification monitoring, to name a few.

## 3. Proposed Approach

### 3.1. Methodology

Figure 1 provides a block diagram of the proposed model. A pretrained VGG-16 network [34] trained for the classification task in ILSVRC [9] is used as feature extractor. The features are then exploited by the regressor which a Fully Connected Neural Network (FCNN) or Recurrent Gated Neural Network (GRU) to output the yaw angle by which the drone should rotate while navigating at fixed forward velocity. The regressor is trained end-to-end and a comparison between both models is provided in Section 5. It should be noted that the convolutional layers used for feature extraction are frozen to avoid extracting features biased towards the synthetic simulation environment. This will help the model to better generalize when deployed for visual path following in real-world settings. Adam [18] is used as an optimizer with learning rate 1e −4 and batch size 64 for 100 epochs. The learning rate is halved every 25 epochs. The loss function is the Mean Squared Error (MSE) between the predicted yaw and the true yaw given as:

$$MSE = \frac{1}{n}\sum_i (y_i - p_i)^2$$

where $n$ is the number of training samples, $y_i$ is the true yaw and $p_i$ is the predicted yaw for sample i.

**Table 1: Information on the paths used in the dataset**

| Path ID | Environment | No. train images | Distance (meters) | Sum of angle change (radians) |
|---|---|---|---|---|
| 1 | Blocks | 6824 | 145.90 | 5.00 |
| 2 | Blocks | 19490 | 239.39 | 4.48 |
| 3 | Landscape | 20993 | 267.22 | 4.23 |
| 4 | Landscape | 32364 | 412.16 | 6.63 |

The true yaw is calculated as the angle between the next waypoint and the drone heading; that is the angle by which the drone deviates from its next waypoint position. The loss function is used to help the network correlate between the visual input, waypoints and the deviation angle. When training on an entire path, the path it is divided into independent waypoint-based segments. It is assumed the waypoints do not overlap and that a model is trained for the entire navigation path.

### 3.2. Control

For simplicity, a fixed drone height of 5 meters and velocity of 1 m/s were used during our experiments. During training phase, the navigation direction at each step is determined as the angle between the y-axis and the vector connecting the current drone position and the position of the next waypoint assuming the path between any two successive waypoints is obstacle-free. In testing, the direction is determined by the CNN and the regressor. We restricted drone motion in all experiments to be only in the direction of its heading.

### 3.3. Flight Path Augmentation

Training only on simulator-generated optimal paths or those acquired during real flights makes the model vulnerable to minor drifts and less adaptable to differences in starting point/heading. For instance, if during testing the drone makes a small drift, a visual feedback slightly different from that used in training will be generated leading to additional drift. Accumulated drift eventually results in total path following mission failure. To mitigate this problem, we introduce the idea of flight path augmentation where many auxiliary paths that are slightly different from the optimal path are generated hence forming a 'navigation envelope' that is subsequently used to train the model. Typically, the auxiliary paths are created by adding noise-based perturbations to both drone position and heading. For instance, the position is perturbed by adding a uniform random value between [-1, 1] meters to the optimal position and the heading yaw is perturbed by adding a uniform random value between [-0.1, 0.1] radians

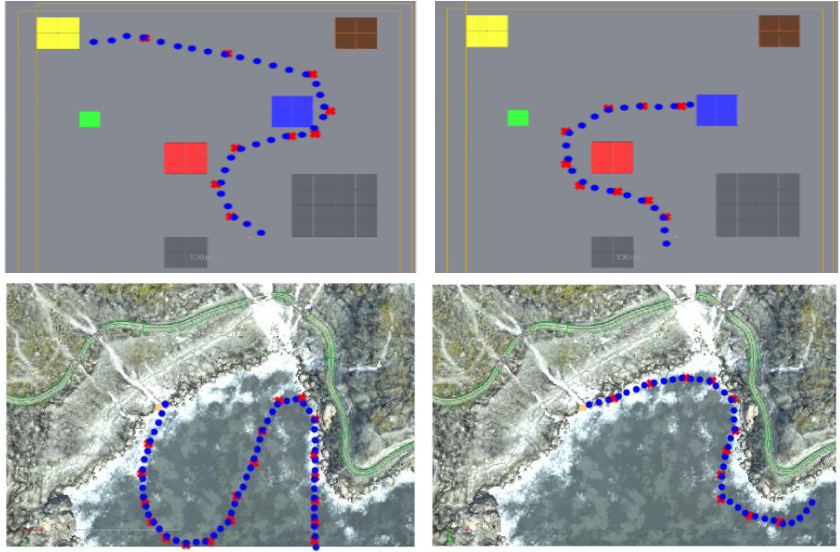

**Figure 3:** Images contain the true paths used in experiments and the paths followed by the drone during testing. The true waypoints are in red and the ones reached by the drone during testing are in blue.

### 3.4. Evaluation Metrics

Two error metrics are used to quantitatively measure the results of different path following models. The first is *Mean Waypoints Minimum Distance* which is the summation of the difference between each waypoint position and the nearest position reached by the drone for the entire path averaged over all waypoints. The second metric is the *Mean Cross Track Distance* [36] which is the shortest distance between the drone position and the next two closest waypoints. A third metric, the Sum of Angle Change, is used to estimate the difficulty of a path in terms of change in navigation angles. In this case, we loop on the path waypoints and calculate the angle between the vector connecting each two successive waypoints and y-axis then accumulate the sum of the difference of these angles.

## 4. Dataset

To collect training dataset, we used the Unreal Engine with the AirSim plugin [32] with physics that provides an API for the drone control and data generation. Two synthetically-generated scenarios were used in the training dataset: Blocks and Landscape. Figure 2 shows sample images from both scenarios. The Blocks scenario represents abstract environment containing cubes with different arrangements and colors. The Landscape environment is a complex scene containing frozen lakes, trees, and mountains. Due to its simplicity, the Blocks scenario was initially used to implement the proposed navigation algorithm while the Landscape scenario was used afterwards to prove the algorithm works in complex realistic scenes.

Two paths are generated in the Blocks environment and two paths in the Landscape environment. As mentioned in Section 3, each path is augmented by adding noise to the optimal shortest path. For each unique path, we generated 100 auxiliary paths with images from the 100 paths used in training. It is worth noting that the model does not have any information about the order of waypoints or the sequence of input images.

Training for each path is done separately and the network learns to follow that path by correlating input images with the yaws required to reach next waypoint. Path-specific training enhances performance since each trained model is conditioned on path starting point and visual feedback. Joint training, on the other hand, could possibly lead to conflicting decisions when the model is faced with similar visual input. Table 1 shows the number of images per each path after augmentation with the 16 jittered paths. Images are of the same size of 512x288 pixels. Distance of the path is measured as the summation of the Euclidean distance between each two successive waypoints.

**Table 2: Evaluation results of four paths using FCNN and GRU with multiple time steps.**

| Path ID | Environment | Random Start | Mean Waypoints Minimum Distance (meters) | | | Mean Cross Track Distance (meters) | | |
|---|---|---|---|---|---|---|---|---|
| | | | FCNN | GRU-2 | GRU-4 | FCNN | GRU-2 | GRU-4 |
| 1 | Blocks | No | 0.92 | 1.2 | 0.94 | 1.62 | 0.93 | 0.92 |
| | | Yes | 0.92 | 1.25 | 1.01 | 1.64 | 0.97 | 0.96 |
| 2 | Blocks | No | 1.12 | 1.57 | 2.51 | 1.55 | 2.63 | 3.90 |
| | | Yes | 1.16 | 1.64 | 2.86 | 1.53 | 2.66 | 4.34 |
| 3 | Landscape | No | 0.86 | 2.62 | 2.77 | 1.07 | 2.85 | 3.44 |
| | | Yes | 0.87 | 2.62 | 2.01 | 1.41 | 2.89 | 3.56 |
| 4 | Landscape | No | 0.82 | 1.28 | - | 1.05 | 1.65 | - |
| | | Yes | 0.83 | 1.29 | - | 1.1 | 1.73 | - |
| Average | - | No | 0.93 | 1.67 | 2.07 | 1.32 | 2.01 | 2.75 |
| | | Yes | 0.945 | 1.7 | 1.9 | 1.42 | 2.06 | 2.95 |

It is worth noting that due to logistical reasons, real drone-acquired data was not included in the training dataset at this stage. In fact, using simulators to develop navigation models that are subsequently tested in real life settings is an emerging trend in development of autonomous navigation models. Recent work has shown that physics engines can be used to learn the dynamics of real-world scenes from images [22, 23] and that models trained in simulation can be generalized to real-world scenarios [41].

## 5. Experiments and Results

In this section, we discuss the experiments conducted to train a model to follow a certain path and deploy it with the minimum possible drift and without using any external positioning system such as GPS. We compare two different regressors: FCNN and GRU with time steps 2 and 4 to test how useful recurrent information is for path following.

We tested our approach in the simulated environments described in Section 4. To make sure that our system is robust and is not just memorizing path images, we added two different kinds of noise during testing. First, random perturbations to flight paths within the navigation envelope. Second, perturbed initial positions and yaws to investigate model behavior in case of different initial deployment conditions.

Table 2 shows the Mean Waypoints Minimum Distance and the Mean Cross Track Distance metrics for the four paths. Results show that FCNN as a regressor achieves lower error in both environments compared to GRU with 2 and 4 timesteps. This result contradicts with results in self-driving car navigation where RNNs provide superior performance in general. The difference between both cases can be attributed to the characteristics of the navigation scene. Autonomous cars utilize road features, which provide largely sequential recurrent information, while navigation. On the other hand, in aerial drone path following, there is no clear visual track to follow unless the drone flying path coincides with a shoreline or distinctive road pattern. Hence, the utilized information is largely the headings to reach the next waypoint on the path. For instance in path 3, the waypoints are near the shoreline however, there is no constant visual contact with the shore line which made GRU performance less efficient compared to FCNN.

It is also shown that that Landscape environment has smaller error compared to the Blocks environment. This is due to the former containing rich visual features that are used for more accurate navigation. The pretrained VGG16 model was trained on images that contain natural images that are more similar to those from the Landscape environment than the Blocks environment. All models were able to follow the path till the end except for GRU with 4 time steps on path 4 as it drifted away on the first sharp turn and couldn't get back on track so we didn't report its error. Figure 3 illustrates several deployment paths when using FCNN as a regressor. It can be seen that our model was able to find the right direction towards the correct paths indicating robust autonomous navigation and improved generalization.

## 6. Conclusions

In this work we presented using visual input for autonomous drone path following that mitigates GPS drawbacks. It has been shown that a CNN combined with a fully connected regressor can successfully predict the steering angles required to move the drone along a pre-determined path. An average of 1.37 meters cross track distance has been achieved across four paths in simulated environments. In future work, we plan to deploy the proposed algorithm on a real drone. We also intend to use Generative Adversarial Networks (GANs) [12, 25] to apply style transfer between synthetic and real training images which is expected to improve system performance and reduce size of training dataset. Other area for future work includes integrating the proposed end-to-end navigation system in UAV middleware [48] as a standalone component and/or within a general framework for target detection and tracking [49][50] that builds on our prior work in these areas.

## References

[1] K. Amer, M. Samy, R. ElHakim, M. Shaker, and M. ElHelw. Convolutional neural network-based deep urban signatures with application to drone localization. In IEEE International Conference on Computer Vision Workshops (ICCVW), 2017 pages 2138–2145. IEEE, 2017.

[2] V. Badrinarayanan, A. Kendall, and R. Cipolla. Segnet: A deep convolutional encoder-decoder architecture for image segmentation. arXiv preprint arXiv:1511.00561, 2015.

[3] M. Bojarski, D. Del Testa, D. Dworakowski, B. Firner, B. Flepp, P. Goyal, L. D. Jackel, M. Monfort, U. Muller, J. Zhang, et al. End to end learning for self-driving cars. arXiv preprint arXiv:1604.07316, 2016.

[4] A. S. Brandao, F. N. Martins, and H. B. Soneguetti. A vision-based line following strategy for an autonomous uav. In Informatics in Control, Automation and Robotics (ICINCO), 2015 12th International Conference on, volume 2, pages 314–319. IEEE, 2015.

[5] D. S. Chaplot, E. Parisotto, and R. Salakhutdinov. Active neural localization. arXiv preprint arXiv:1801.08214, 2018.

[6] Z. Chen and S. T. Birchfield. Qualitative visionbased path following. IEEE Transactions on Robotics, 25(3):749–754, 2009.

[7] R. Clark, S. Wang, H. Wen, A. Markham, and N. Trigoni. Vinet: Visual-inertial odometry as a sequence-to-sequence learning problem. In AAAI, pages 3995–4001, 2017.

[8] D. H. De Mel, K. A. Stol, J. A. Mills, and B. R. Eastwood. Vision-based object path following on a quadcopter for gps-denied environments. In Unmanned Aircraft Systems (ICUAS), 2017 International Conference on, pages 456–461. IEEE, 2017.

[9] J. Deng, W. Dong, R. Socher, L.-J. Li, K. Li, and L. Fei-Fei. Imagenet: A large-scale hierarchical image database. In Computer Vision and Pattern Recognition, 2009. CVPR 2009. IEEE Conference on, pages 248–255. Ieee, 2009.

[10] R. Girshick. Fast r-cnn. In Proceedings of the IEEE international conference on computer vision, pages 1440–1448, 2015.

[11] A. Giusti, J. Guzzi, D. C. Ciresan, F.-L. He, J. P. Rodríguez, F. Fontana, M. Faessler, C. Forster, J. Schmidhuber, G. Di Caro, et al. A machine learning approach to visual perception of forest trails for mobile robots. IEEE Robotics and Automation Letters, 1(2):661–667, 2016.

[12] I. Goodfellow, J. Pouget-Abadie, M. Mirza, B. Xu, D. Warde-Farley, S. Ozair, A. Courville, and Y. Bengio. Generative adversarial nets. In Advances in neural information processing systems, pages 2672–2680, 2014.

[13] A. Hernandez, C. Copot, R. De Keyser, T. Vlas, and I. Nascu. Identification and path following control of an ar. drone quadrotor. In System Theory, Control and Computing (ICSTCC), 2013 17th International Conference, pages 583–588. IEEE, 2013.

[14] B. K. Horn and B. G. Schunck. Determining optical flow. Artificial intelligence, 17(1-3):185–203, 1981.

[15] K. Kelchtermans and T. Tuytelaars. How hard is it to cross the room?–training (recurrent) neural networks to steer a uav. arXiv preprint arXiv:1702.07600, 2017.

[16] A. Kendall, M. Grimes, and R. Cipolla. Posenet: A convolutional network for real-time 6-dof camera relocalization. In Proceedings of the IEEE international conference on computer vision, pages 2938–2946, 2015.

[17] D. K. Kim and T. Chen. Deep neural network for realtime autonomous indoor navigation. arXiv preprint arXiv:1511.04668, 2015.

[18] D. P. Kingma and J. Ba. Adam: A method for stochastic optimization. arXiv preprint arXiv:1412.6980, 2014.

[19] A. Krizhevsky, I. Sutskever, and G. E. Hinton. Imagenet classification with deep convolutional neural

networks. In Advances in neural information processing systems, pages 1097–1105, 2012.

[20] B. D. Lucas, T. Kanade, et al. An iterative image registration technique with an application to stereo vision. 1981.

[21] I. Melekhov, J. Ylioinas, J. Kannala, and E. Rahtu. Relative camera pose estimation using convolutional neural networks. In International Conference on Advanced Concepts for Intelligent Vision Systems, pages 675–687. Springer, 2017.

[22] R. Mottaghi, H. Bagherinezhad, M. Rastegari, and A. Farhadi. Newtonian scene understanding: Unfolding the dynamics of objects in static images. In Proceedings of the IEEE Conference on Computer Vision and Pattern Recognition, pages 3521–3529, 2016.

[23] R. Mottaghi, M. Rastegari, A. Gupta, and A. Farhadi. what happens if... learning to predict the effect of forces in images. In European Conference on Computer Vision, pages 269–285. Springer, 2016.

[24] T. Nguyen, G. K. Mann, and R. G. Gosine. Visionbased qualitative path-following control of quadrotor aerial vehicle. In Unmanned Aircraft Systems (ICUAS), 2014 International Conference on, pages 412–417. IEEE, 2014.

[25] A. Radford, L. Metz, and S. Chintala. Unsupervised representation learning with deep convolutional generative adversarial networks. arXiv preprint arXiv:1511.06434, 2015.

[26] J. Redmon, S. Divvala, R. Girshick, and A. Farhadi. You only look once: Unified, real-time object detection. In Proceedings of the IEEE conference on computer vision and pattern recognition, pages 779–788, 2016.

[27] O. Ronneberger, P. Fischer, and T. Brox. U-net: Convolutional networks for biomedical image segmentation. In International Conference on Medical image computing and computer-assisted intervention, pages 234–241. Springer, 2015.

[28] S. Ross, G. Gordon, and D. Bagnell. A reduction of imitation learning and structured prediction to noregret online learning. In Proceedings of the fourteenth international conference on artificial intelligence and statistics, pages 627–635, 2011.

[29] S. Ross, N. Melik-Barkhudarov, K. S. Shankar, A. Wendel, D. Dey, J. A. Bagnell, and M. Hebert. Learning monocular reactive uav control in cluttered natural environments. arXiv preprint arXiv:1211.1690, 2012.

[30] L. V. Santana, A. S. Brandao, M. Sarcinelli-Filho, and R. Carelli. A trajectory tracking and 3d positioning controller for the ar. drone quadrotor. In Unmanned Aircraft Systems (ICUAS), 2014 International Conference on, pages 756–767. IEEE, 2014.

[31] P. Sermanet, D. Eigen, X. Zhang, M. Mathieu, R. Fergus, and Y. LeCun. Overfeat: Integrated recognition, localization and detection using convolutional networks. arXiv preprint arXiv:1312.6229, 2013.

[32] S. Shah, D. Dey, C. Lovett, and A. Kapoor. Airsim: High-fidelity visual and physical simulation for autonomous vehicles. In Field and service robotics, pages 621–635. Springer, 2018.

[33] J. Shi and C. Tomasi. Good features to track. Technical report, Cornell University, 1993. [34] K. Simonyan and A. Zisserman. Very deep convolutional networks for large-scale image recognition. arXiv preprint arXiv:1409.1556, 2014.

[35] N. Smolyanskiy, A. Kamenev, J. Smith, and S. Birchfield. Toward low-flying autonomous mav trail navigation using deep neural networks for environmental awareness. arXiv preprint arXiv:1705.02550, 2017.

[36] P. Sujit, S. Saripalli, and J. B. Sousa. An evaluation of uav path following algorithms. In Control Conference (ECC), 2013 European, pages 3332–3337. IEEE, 2013.

[37] N. O. Tippenhauer, C. Popper, K. B. Rasmussen, and ¨ S. Capkun. On the requirements for successful gps spoofing attacks. In Proceedings of the 18th ACM conference on Computer and communications security, pages 75–86. ACM, 2011.

[38] S. Wang, R. Clark, H. Wen, and N. Trigoni. Deepvo: Towards end-to-end visual odometry with deep recurrent convolutional neural networks. In Robotics and Automation (ICRA), 2017 IEEE International Conference on, pages 2043–2050. IEEE, 2017.

[39] J. S. Warner and R. G. Johnston. Gps spoofing countermeasures. Homeland Security Journal, 25(2):19–27, 2003.

[40] T. Zhang, G. Kahn, S. Levine, and P. Abbeel. Learning deep control policies for autonomous aerial vehicles with mpc-guided policy search. arXiv preprint arXiv:1509.06791, 2015.

[41] Y. Zhu, R. Mottaghi, E. Kolve, J. J. Lim, A. Gupta, L. Fei-Fei, and A. Farhadi. Target-driven visual navigation in

indoor scenes using deep reinforcement learning. In Robotics and Automation (ICRA), 2017 IEEE International Conference on, pages 3357–3364. IEEE, 2017.

[42] Amazon drone delivery system patent: https://patents.google.com/patent/US9573684B2/en

[43] UAV drug delivery white paper: http://www.villagereach.org/wp-content/uploads/2018/02/JSI-UAV-Report.pdf

[44] Uber Air Taxi white paper: www.uber.com/elevate.pdf

[45] https://blog.ferrovial.com/en/2017/06/drones-for-environmental-monitoring/

[46] https://www.techtimes.com/articles/170974/20160722/facebooks-aquila-soars-in-test-flight-solar-powered-drone-will-provide-internet-access-to-everyone.htm

[47] Cho, Kyunghyun, et al. "Learning phrase representations using RNN encoder-decoder for statistical machine translation." arXiv preprint arXiv:1406.1078 (2014).

[48] A. El-Sayed and M. ElHelw, "Distributed component-based framework for unmanned air vehicle systems", Proceedings of the IEEE International Conference on Information and Automation, 2012.

[49] A. Salaheldin, S. Maher, M. ElHelw. "Robust real-time tracking with diverse ensembles and random projections", Proceedings of the IEEE International Conference on Computer Vision (ICCV) Workshops, 2013, pp. 112-120.

[50] M. Siam, M. Elhelw. "Enhanced target tracking in UAV imagery with PN learning and structural constraints", Proceedings of the IEEE International Conference on Computer Vision (ICCV) Workshops, 2013, pp. 586-593.